\documentclass[11 pt, onecolumn]{article}
\usepackage[margin=1.1in]{geometry}

\usepackage{amsmath}
\usepackage{bm}	
\usepackage{graphicx}

\usepackage{graphicx}
\usepackage{multirow}
\usepackage{mdwlist}

\usepackage{threeparttable}
\DeclareMathOperator*{\argmax}{arg\,max}
\usepackage{amsfonts}
\usepackage{amsmath}
\usepackage{hyperref}
\usepackage{bbm}
\usepackage{float}
\usepackage{caption}
\usepackage{subcaption}
\usepackage{xcolor}
\usepackage{mathtools}
\usepackage{tikz}

\usetikzlibrary{shapes.geometric,arrows}
\usepackage{enumerate}
\usepackage{multicol}
\usepackage{stackrel}

\usepackage{subcaption}
\usepackage{amssymb}
\usepackage[utf8]{inputenc}
\usepackage{booktabs,lipsum}
\usepackage{hyperref}
\usepackage{algorithm}
\usepackage{algpseudocode}
\usepackage{caption}

\DeclareMathAlphabet{\mathcal}{OMS}{cmsy}{m}{n}

\usepackage{graphicx}
\usepackage{epstopdf}

\begin{document}
\title{\LARGE \bf Dynamic Origin-Destination Matrix Prediction with Line Graph Neural Networks and Kalman Filter}

\author{Xi Xiong, Kaan Ozbay, Li Jin, and Chen Feng
\thanks{This work was supported in part by NYU Tandon School of Engineering and C2SMART Department of Transportation Center. The authors appreciate the discussion with Professor Bekir Bartin.}
\thanks{X. Xiong, K. Ozbay, L. Jin, and C. Feng are with the Department of Civil and Urban Engineering, and C. Feng is joint with Department of Mechanical and Aerospace Engineering, New York University Tandon School of Engineering, Brooklyn, NY, USA, emails: xi.xiong@nyu.edu, kaan.ozbay@nyu.edu, lijin@nyu.edu, cfeng@nyu.edu}%
}
\newcommand*{\QEDA}{\hfill\ensuremath{\blacksquare}}%

\maketitle

\begin{abstract}
Modern intelligent transportation systems provide data that allow real-time dynamic demand prediction, which is essential for planning and operations. The main challenge of prediction of dynamic  Origin-Destination (O-D) demand matrices is that demands cannot be directly measured by traffic sensors; instead, they have to be inferred from aggregate traffic flow data on traffic links. Specifically, spatial correlation, congestion and time dependent factors need to be considered in general transportation networks. In this paper we propose a novel O-D prediction framework combining heterogeneous prediction in graph neural networks and Kalman filter to recognize spatial and temporal patterns simultaneously. The underlying road network topology is converted into a corresponding line graph in the newly designed Fusion Line Graph Convolutional Networks (FL-GCNs), which provide a general framework of predicting spatial-temporal O-D flows from link information. Data from New Jersey Turnpike network are used to evaluate the proposed model. The results show that our proposed approach yields the best performance under various prediction scenarios. In addition, the advantage of combining deep neural networks and Kalman filter is demonstrated.
\end{abstract}


\section{Introduction}
Traffic demand is typically characterized by an Origin-Destination (O-D) matrix, in which the elements denote the number of trips between O-D pairs during a certain time interval. O-D flows are fundamental prerequisites for transportation analysis, and can provide trip patterns among geological zones, which can reflect traffic and economic activities. Reliable prediction of O-D flows can improve planning and operations in real-time traffic management \cite{djahel2015communications,de2017traffic,jin2018throughput}.
Furthermore, with the development of connected and autonomous vehicles (CAVs), dynamic O-D information can facilitate the process of vehicle assignment and route choice, which can improve the efficiency of intelligent transportation system (ITS). However, O-D matrices are not directly accessible by traffic sensors; instead, they have been estimated by household surveys, which are expensive and time consuming. Alternatively, O-D matrices have been inferred from link counts in the surveillance system, which is notoriously difficult to predict spatial-temporal correlated demands due to the complexity of traffic networks.

Consider a transportation network with multiple O-D pairs. Link flows can be obtained from traffic sensors such as loop inductors in the surveillance system, which can provide traffic count, speed and incident information. Corresponding algorithms can be designed to forecast multi-step traffic demands by combing historical O-D flows and link flows \cite{zhang2019multistep}. The network topology is essential during the prediction process, which can distinguish the O-D forecasting problem in a given transportation network from other inference problems. The network topology operates as the prior knowledge to assist the data-driven prediction. Generally, the network topology is denoted by a matrix that presents the link connection, node connection or implicit transformation between links and nodes \cite{ashok1996estimation}. Obviously, the objective in this problem is to minimize the prediction errors in corresponding O-D matrices.

In this paper, we propose a framework that combines graph neural networks and Kalman filter to forecast O-D flows. Graph neural networks are shown to be effective in processing data with the specific network topology, which is denoted by an adjacency matrix. Kalman filter is a classical model including prediction and update steps to minimize prediction uncertainties \cite{ashok2000alternative}. Since graph neural networks and Kalman filter utilize different topology matrices and optimization mechanisms, a mixing parameter is used to balance heterogeneous prediction in the two methods. In addition, we design the novel Fusion Line Graph Convolutional Networks (FL-GCNs) including link graph convolution and node graph convolution to predict O-D flows. The proposed networks provide a general deep learning framework to deal with problems related to spatial-temporal mapping from links to nodes. Real data in New Jersey Turnpike are used to evaluate our model. The results show that the combining model yields the best performance. The effect of balancing graph neural networks and Kalman filter, and the converged weights in deep neural networks are revealed. The main contributions of this paper are:

(a). We propose the structure that combines graph neural networks with Kalman filter to predict spatial-temporal O-D flows. Since different topology matrices are incorporated into graph neural networks and Kalman filter, a mixing parameter is used to balance outputs in two models, which can improve prediction accuracy and robustness under different steps.

(b). We design the novel Fusion Line Graph Convolutional Networks (FL-GCNs) including link graph convolution and node graph convolution. This structure can be applied to problems related to spatial-temporal aggregation from link information to node information.

(c). We validate our approach by real-world case study in New Jersey Turnpike. The results show that the combining method yields the best performance in forecasting O-D flows. We then investigate the characteristics of both methods. In addition, the effect of balancing both methods and converged weights in deep neural networks are analyzed.

The rest of this paper is organized as follows. We first review related work on O-D flow estimation/prediction. Next, we elaborate on our proposed model for O-D demand prediction. Then, we use real traffic data to evaluate our approach and show our results and analysis. Finally, we summarize the conclusions and propose several directions for future work.

\section{Literature Review}
\label{Section: review}
Prediction of O-D flows has been studied for decades. Several models have been proposed to solve the problem. Gravity Model is a widely used approach to tackle static O-D prediction problems. However, its effectiveness is limited due to highly dynamic and nonlinear features of transportation flows that cannot be captured by its underlying mathematical structure. Statistical models, such as Generalized Least Squares (GLS), Maximum Likelihood (ML) estimation, and Bayesian methods are widely used to solve the O-D estimation and prediction problems. The objective of GLS is to minimize the difference between estimated flows and observed flows \cite{cascetta1984estimation, bell1991estimation}. ML estimators are obtained by maximizing the likelihood of observed flows conditional on estimated O-D flows \cite{spiess1987maximum}. In the framework of Bayesian approach, posterior probability is calculated by combining prior probability expressed by O-D flows and link flow likelihood conditional on estimated O-D flows. The Bayesian solution \cite{maher1983inferences, cascetta1988unified} is to find O-D flows that would maximize the posterior probability.

Advanced models take spatial and temporal effects into consideration. For the spatial part, the key problem is the mapping from O-D flows to link flows. Assignment matrices are usually used to represent this relationship. There are two steps from O-D flows to link flows. The first step is the mapping from O-D flows to path flows. Route choice behaviors are considered in this stage. The second step is from path flows to link flows. When the path is not congested, the mapping of path flows to link flows is given by a link-path incidence matrix. In general transportation networks with congestion, User Equilibrium (UE) is usually used to characterize route choice behaviors. The bi-level O-D estimation method incorporates the UE assumption \cite{yang1995heuristic, maher2001bi}. The upper level is to minimize the difference between estimated and observed flows, and the lower level is to determine a flow pattern that satisfies user equilibrium conditions. Another method that incorporates UE is the Path Flow Estimator (PFE) \cite{sherali1994linear, sherali2001estimation}. The object of PFE is to find the optimized path flows. The estimated O-D flows are calculated by adding up flows on all paths connecting respective O-D pair. For the temporal part, Okutani \cite{okutani1987kalman} and Ashok and Ben-Akiva \cite{ashok2002estimation} used Kalman filter to represent the dynamic transition between consequent O-D flows. The transition equation uses an auto-regressive model to predict future O-D flows based on prior ones. The measurement equation denotes the relationship between O-D flows and link flows to capture the topology of transportation networks. In this paper, we use Kalman filter as a benchmark to compare the performance of our proposed approach.

In recent years, deep neural networks have shown to be effective in approximating nonlinear features in classification, regression and control problems \cite{lecun2015deep, silver2018general}. Up to date, supervised learning plays a major part in the field of deep learning. Convolutional Neural Networks (CNNs) and Recurrent Neural Networks (RNNs) are effective in recognizing spatial and temporal patterns respectively \cite{krizhevsky2012imagenet, xingjian2015convolutional}. Since transportation networks are denoted by nodes and arcs, graph-structured data appear frequently in this domain. Graph Neural Networks (GNNs) are shown to be effective in dealing with graph-structured data \cite{battaglia2018relational}. Graph Convolutional Networks (GCNs) \cite{kipf2016semi} utilize the adjacency matrix to represent node connections. Incorporating network topology into deep neural networks can accelerate convergence and improve prediction performance. Gated Graph Neural Networks (GGNNs) \cite{li2015gated} have shown outstanding performance in time series graph tasks. Yu et al. \cite{yu2017spatio} extended GCN to time series structure and proposed an integrated framework for spatial-temporal graph traffic forecasting. Although this structure can represent the evolution of node information, the information flow from arcs to nodes cannot be reflected. Chen et al. \cite{chen2018supervised} proposed the Line Graph Neural Networks (LGNNs) to solve the problem. However, this structure requires extensive information exchange between nodes and arcs, which can increase computing burden in practice. 
\section{Methodology} \label{Section: model}
In this section, we would present the framework that combines graph neural networks and Kalman filter to predict Origin-Destination (O-D) flows. Preliminary definitions and notations throughout this paper are firstly introduced. Then we elaborate on our proposed methodology including graph neural networks and Kalman filter.

\subsection{Preliminary Definitions}
Consider a directed graph $\mathcal{G} = (\mathcal{V}, \mathcal{E})$ that includes a set of nodes $\mathcal{V}$ and a set of  links $\mathcal{E}$. The network
consists of $n_{d}$ nodes, $n_{k}$ links and $n_{od} = n_d (n_{d}-1)$ O-D pairs. We assume that $n_l$ of the $n_{k}$ links are equipped with traffic sensors.

During an analysis period divided into equal intervals $h=1,2,3,\ldots$, $x_{rh}$ represents the number of vehicles between the $r^{th}$ O-D pair departed in interval $h$. The number of traffic counts at detector $l$ during interval $h$ is denoted by $y_{lh}$. We use $\mathbf{x}_h$ to denote corresponding $(n_d \times (n_d-1))$ vector of all O-D pairs, and use $\mathbf{y}_h$ to represent corresponding $(n_{l} \times 1)$ vector of all link flows. In addition, $\mathbf{x}{_h^H}$ and $\mathbf{y}{_h^H}$ are used to denote corresponding historical O-D flows and link flows, which  typically are the counts in interval $h$ during previous days. To fit the structure of neural networks and incorporate time series information, we integrate real-time link flows with historical link flows, and use the $(n_{l} \times 2t)$ vector $\mathbf{Z}_{h-1}$ to denote the integrated link flows as:
\begin{align}
    \mathbf{Z}_{h-1} = & \big[\mathbf{y}_{h-1} \quad \mathbf{y}_{h-2} \quad \ldots \quad \mathbf{y}_{h-t} \quad \mathbf{y}{_{h-1}^H} \quad \mathbf{y}{_{h-2}^H} \quad \ldots  \quad 
     \mathbf{y}{_{h-t}^H} \big],
\end{align}
where $t$ denotes the time series information, i.e., the number of prior intervals in link flows. Then the task of $k^{th}$ step O-D prediction problem can be formulated as
\begin{align}
 \mathbf{\hat{x}}_{h+k-1}  
 & = \argmax_{\mathbf{x}_{h+k-1}} \Pr \big( \mathbf{x}_{h+k-1} | \mathbf{Z}_{h-1} ; \mathbf{x}{_h^H}, \mathbf{x}{_{h-1}^H},\mathbf{x}{_{h-2}^H}  \ldots, \mathbf{x}{_{h-m}^H}; \mathcal{G} \big),
\end{align}
where $m$ is the number of prior intervals in O-D flows, and the prediction of $\mathbf{x}_{h+k-1}$ is denoted by $\mathbf{\hat{x}}_{h+k-1}$. $\Pr (\cdot | \cdot)$ denotes the function of conditional probability based on historical data and network topology. Notations used in this paper are introduced in Table \ref{table:notations}.

\begin{table}[hbt]
  \begin{center}
  \caption{Illustration of notations.}\label{table:notations}
  \begin{tabular}{cc}
    \hline
    $n_{d}$              & number of nodes             \\
    $n_{k}$              & number of links             \\
    $n_{l}$              & number of links  equipped with sensors            \\
    $n_{od}$             & number of O-D pairs        \\
    $x_{rh}$     & $r^{th}$ O-D flow departed in interval $h$        \\
    $y_{lh}$             & link counts at detector $l$ during interval $h$             \\
    $\mathbf{x}_h$            & $(n_d \times (n_d-1))$ vector of O-D pairs in interval $h$     \\
    $\mathbf{y}_h$             & $(n_{l} \times 1)$ vector of link counts in interval $h$       \\
    $\mathbf{x}{_h^H}$             & historical $h$ interval O-D flows        \\
    $\mathbf{y}{_h^H}$             & historical $h$ interval link flows        \\
    $\mathbf{\hat{x}}_h$            & the prediction of O-D matrix $\mathbf{x}_h$        \\
    $\mathbf{Z}_{h-1}$             & integration of link flows and historical link flows      \\
    $\mathbf{A}_L$             & link adjacency matrix in neural networks       \\
    $\mathbf{A}_N$             & node adjacency matrix in neural networks       \\
    $\mathbf{P}$               & incidence matrix in neural networks       \\
    $\mathbf{\widetilde{x}}_0$             & initial state vector in Kalman filter        \\
    $\mathbf{p}_0$             & initial covariance in Kalman filter        \\
    $\mathbf{f}$             & transition matrix in Kalman filter        \\
    $\mathbf{w}$             & transition error in Kalman filter       \\
    $\mathbf{a}$             & assignment matrix in Kalman filter       \\
    $\mathbf{v}$             & measurement error in Kalman filter        \\
    $\partial{\mathbf{x_h}}$             & O-D matrix deviations\\
    \hline
    \end{tabular}
    \end{center}
\end{table}

\subsection{Model Framework}
\begin{figure}[hbt]
  \centering
  \includegraphics[width=0.75\textwidth, trim=70 60 60 80,clip]{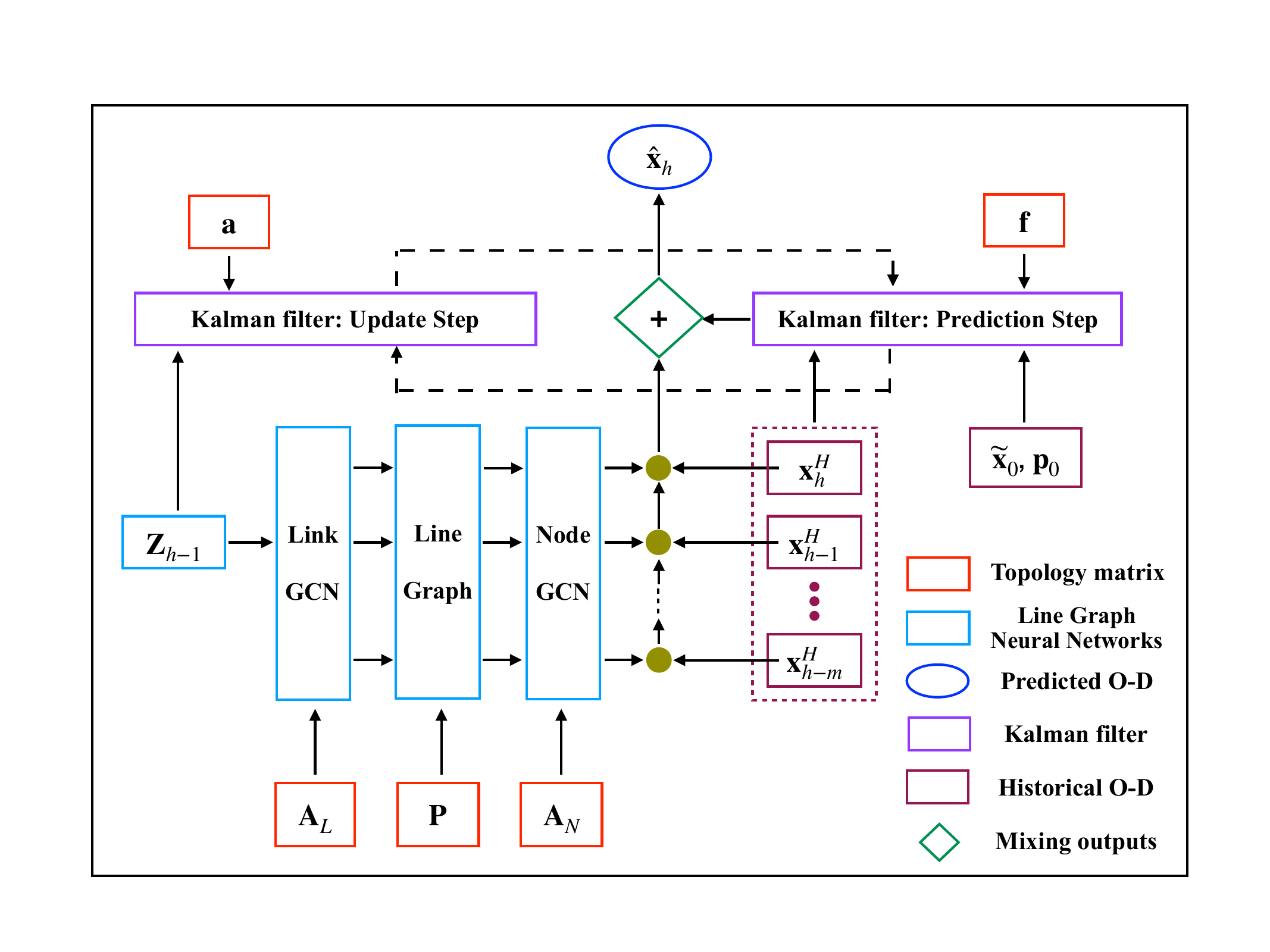}
  \caption{The structure of our model.}\label{fig:structure}
\end{figure}

The framework of our proposed model (Figure \ref{fig:structure}) consists of two major parts: graph neural networks and Kalman filter. Inputs for our model are integrated link flows and historical O-D flows, and the outputs are the predicted O-D matrices.

We design the novel Line Graph Convolutional Networks (L-GCNs) to convert link flows into O-D matrices through a series of link graph convolution and node graph convolution. Historical O-D flows, which incorporate prior O-D demand patterns, are fused into L-GCNs to construct the Fusion Line Graph Convolutional Networks (FL-GCNs). In the Kalman filter section, we first use historical link flows and O-D flows to construct the transition matrix and the assignment matrix. The O-D matrices are predicted through prediction and update steps. During each update step, we use real-time link flows to minimize prediction covariance. Then the corrected O-D estimation is used in the prediction step to output the O-D matrix in the next interval.

Deep neural networks utilize historical data to recognize dynamic spatial-temporal patterns. In comparison, Kalman filter uses current link flows to update estimation uncertainty step by step. Furthermore, FL-GCN and Kalman filter utilize different topology matrices. We then use a mixing parameter to balance heterogeneous prediction in two methods.

\subsection{Line Graph Convolutional Networks (L-GCNs)}

\subsubsection{Graph Convolutional Networks (GCNs)}
The GCN proposed by Kipf and Welling \cite{kipf2016semi} is shown to be efficient in learning on graph-structured data. In our O-D prediction problem, the GCN incorporating the traffic topology into neural networks is firstly used to denote the evolution of link flows.

The $(n_{l} \times n_{l})$ adjacent matrix $\mathbf{A}_L$ is used to represent the link connections. Then we use spectral convolution to predict $\mathbf{\hat{y}}_{h}$, which denotes the predicted link flows in interval $h$. In this case, the $n_{l}$ link detectors are considered as hypothetical nodes by the transformation of line graph in Figure \ref{fig:line_graph_transformation}. Features in each hypothetical node include real-time link flows and historical link flows in vector $\mathbf{Z}_{h-1}$. The number of features is $2t$.

The spectral operation on graphs is defined as the multiplication of a signal $\mathbf{z} \in \mathbb{R}^{n_l}$ (a scalar for each node) with a filter $\mathbf{g}_{\theta}=$ diag $(\theta)$ parameterized by $\theta \in \mathbb{R}^{n_l}$ in the Fourier domain:
\begin{align} \label{Equation:eigenvectors}
\mathbf{g}_{\theta} \star \mathbf{z} = \mathbf{U} \mathbf{g}_{\theta} (\bm{\Lambda}) \mathbf{U}^T \mathbf{z},
\end{align}
where $\mathbf{U}$ is the matrix of eigenvectors of the graph Laplacian $\mathbf{L} = \mathbf{I}_{n_l} - \mathbf{D}^{- \frac{1}{2}} \mathbf{A}_L \mathbf{D}^{- \frac{1}{2}} = \mathbf{U} \bm{\Lambda} \mathbf{U}^T$, with a diagonal matrix of its eigenvalues $\bm{\Lambda}$. $\mathbf{D}$ is the degree matrix and $\mathbf{I}_{n_l}$ is the identity matrix. Equation \eqref{Equation:eigenvectors} can be approximated by
$\mathbf{g}_{\theta} \star \mathbf{z} \approx \theta \left(\mathbf{I}_{n_l} + \mathbf{D}^{- \frac{1}{2}} \mathbf{A}_L \mathbf{D}^{- \frac{1}{2}} \right) \mathbf{z}$.
Kipf and Welling \cite{kipf2016semi} introduced the renormalization trick and replaced $\left( \mathbf{I}_{n_l} + \mathbf{D}^{- \frac{1}{2}} \mathbf{A}_L \mathbf{D}^{- \frac{1}{2}} \right)$ with $\mathbf{\widetilde{D}}^{- \frac{1}{2}} \mathbf{\widetilde{A}}_L \mathbf{\widetilde{D}}^{- \frac{1}{2}}$, in which $\mathbf{\widetilde{A}}_L = \mathbf{A}_L + \mathbf{I}_{n_l}$ and $\mathbf{\widetilde{D}}_{ii} = \sum_{j}^{} \mathbf{\widetilde{A}}_{L_{ij}}$. In our case, 
$\mathbf{\widetilde{D}}^{- \frac{1}{2}} \mathbf{\widetilde{A}}_L \mathbf{\widetilde{D}}^{- \frac{1}{2}}$ is replaced with 
the Random Walk Laplacian matrix $\mathbf{\hat{A}}_L = \mathbf{\widetilde{D}}^{-1} \mathbf{\widetilde{A}}_L$ to simplify the expression. Then the signal $ \mathbf{z} \in \mathbb{R}^{n_l}$ can be extended to $\mathbf{Z} \in \mathbb{R}^{n_l \times 2t}$ with $2t$ input features, and the convolution operation can be generalized by:
\begin{align}
\mathbf{\hat{y}} = \mathbf{\hat{A}}_L \mathbf{Z} \bm{\Theta},
\end{align}
where $\bm{\Theta} \in \mathbb{R}^{2t \times 1}$ is a matrix of filter parameters, and $\mathbf{\hat{y}}$ is the $(n_l \times 1)$ convolved matrix. We consider a two-layer GCN as:
\begin{align}
\label{Equation: GCN}
\mathbf{\hat{y}}^1_h = \rho \left(\mathbf{\hat{A}}_L \sigma \left(\mathbf{\hat{A}}_L \mathbf{Z}_{h-1} \mathbf{w}_0 + \mathbf{b}_0 \right) \mathbf{w}_1 + \mathbf{b}_1 \right),
\end{align}
where $\mathbf{\hat{y}}^1_h$ is the convolved link flows using link GCN, $\rho(\cdot)$ and $\sigma(\cdot)$ are activation functions, and $\mathbf{w}_0$, $\mathbf{w}_1$, $\mathbf{b}_0$, $\mathbf{b}_1$ represent parameters in each layer.

The modified adjacency matrix $\mathbf{\hat{A}}_L$, which has the same function as the assignment matrix in Equation \eqref{Equation1b}, denotes the topology information, and would accelerate the convergence of deep neural networks.

\subsubsection{Newly designed Line Graph Convolutional Networks (L-GCNs)}
In our O-D prediction problem, we need to represent the evolution from links $\mathcal{E}$ to nodes $\mathcal{V}$. In this section, we would transform the original directed graph $\mathcal{G} = (\mathcal{V}, \mathcal{E})$ into corresponding line graph and show the structure of L-GCN.

\begin{figure}[!ht]
  \centering
  \begin{subfigure}[b]{1.0\textwidth}
  \centering
  \includegraphics[width=0.63\textwidth, trim=40 130 40 160,clip]{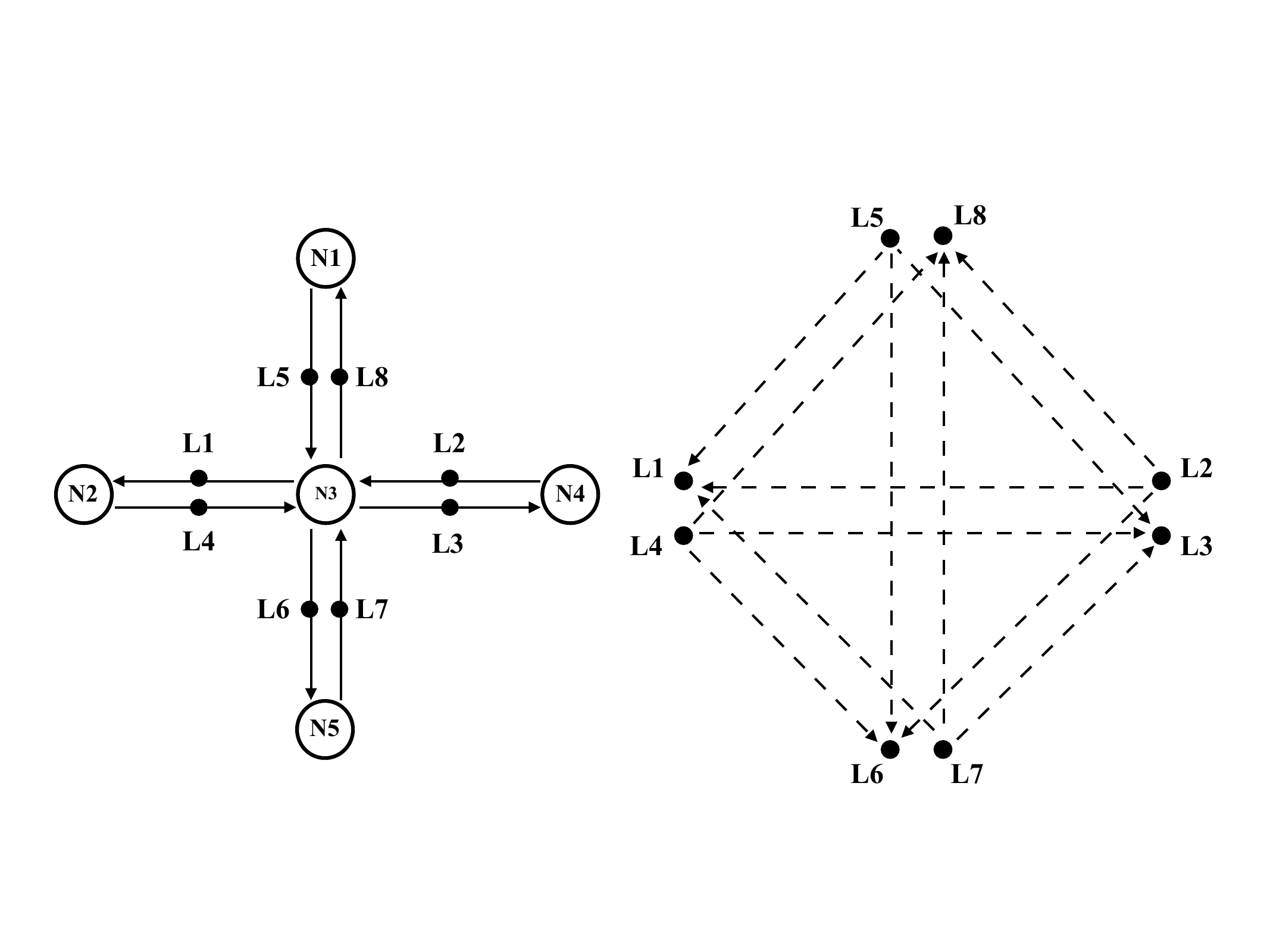}
  \caption{Transformation of line graph}
  \label{fig:line_graph_transformation}
  \end{subfigure}
  \hfill
  \begin{subfigure}[b]{1.0\textwidth}
  \centering
  \includegraphics[width=0.63\textwidth, trim=50 150 130 220,clip]{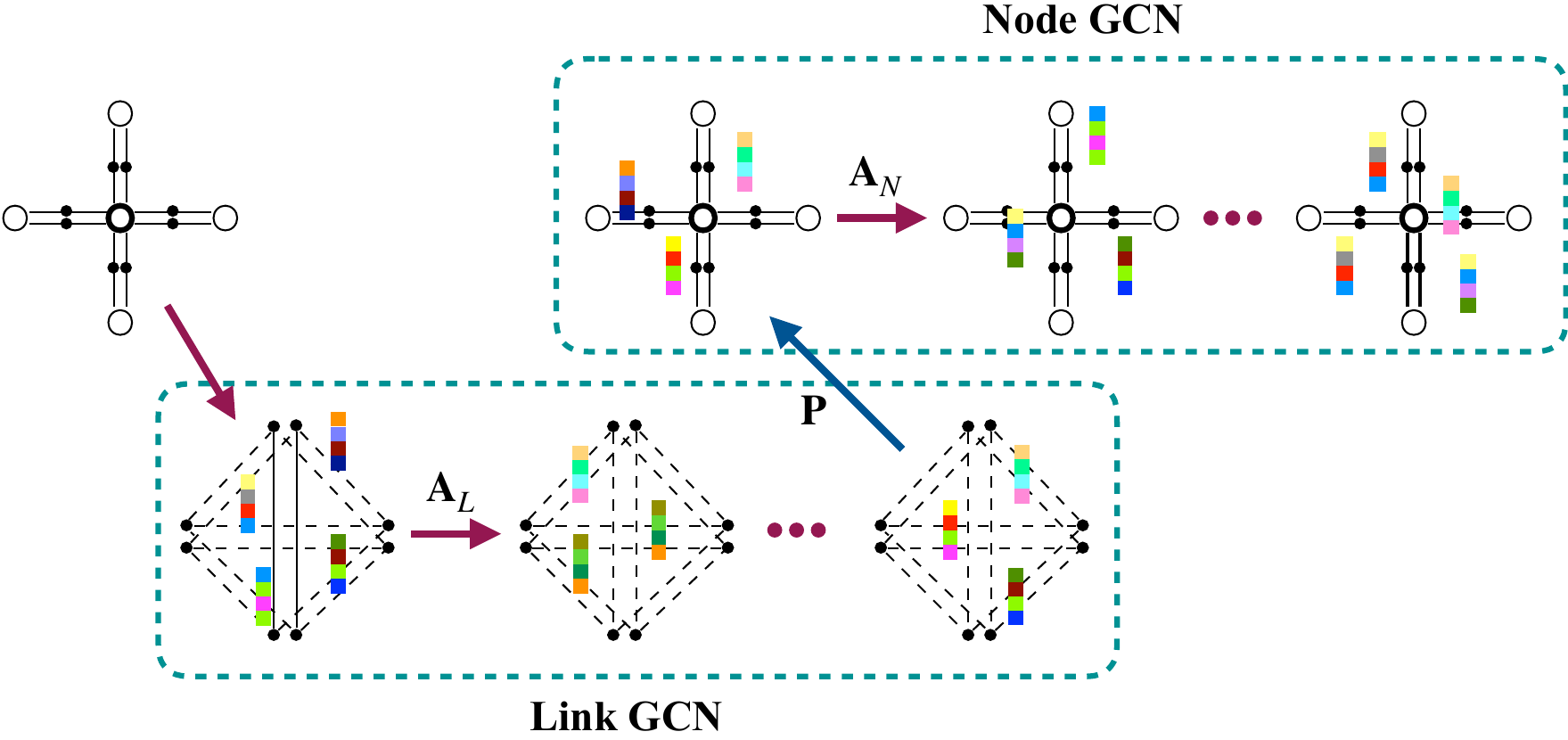}
  \caption{Matrix iteration in L-GCN}
  \label{fig:matrix_iteration}
  \end{subfigure}
  \caption{Line graph transformation and graph convolution in links and nodes}\label{fig:line_graph}
\end{figure}

Consider a directed graph such as that in Figure \ref{fig:line_graph_transformation}. The original graph can be transformed into corresponding line graph. Let $\mathcal{L}(\mathcal{G})$ represent this operation.
The line graph represents adjacent relationship between edges of $\mathcal{G}$. An incidence matrix $\mathbf{P}$ is used to represent the aggregation from links to nodes (Figure \ref{fig:matrix_iteration}). Consider node $i$ and node $j$ in the graph $\mathcal{G}$. When $P_{ij}$ represents the link starting from node $i$ and the outflow of node $i$, the value in the incidence matrix is $1$. When $P_{ij}$ denotes the inflow of node $i$, the corresponding value is $-1$. If there is no connection between a link and a node, the value in the incidence matrix is $0$.

In Figure \ref{fig:line_graph}, we would obtain the node adjacency matrix $\mathbf{A}_N$, the link adjacency matrix $\mathbf{A}_L$ and the incidence matrix $\mathbf{P}$. Chen et al. \cite{chen2018supervised} proposed the Line Graph Neural Networks (LGNNs) that considered the interaction between nodes and links. LGNN includes the evolution of original graph $\mathcal{G}$, the evolution of line graph $\mathcal{L}(\mathcal{G})$, and the interaction between them. However, this structure requires extensive information exchange between nodes and links, which would increase computing burden in practice. In our approach, we relax some connections in LGNN and only consider the link graph convolution and node graph convolution (Figure \ref{fig:matrix_iteration}). The evolution of link flows is denoted by the GCN, the aggregation of link flows is represented by the incidence matrix $\mathbf{P}$, and the node graph convolution is also given by the GCN. The proposed L-GCN is shown as follows:
\begin{align}
\mathbf{x}{_h^L} & = \mathbf{\hat{A}}_N \phi \left( \mathbf{P}  \mathbf{\hat{y}}^1_h \mathbf{w}_2 + \mathbf{b}_2 \right) \mathbf{w}_3 +                               \mathbf{b}_3, 
\end{align}
where $\mathbf{x}{_h^L}$ is the predicted O-D matrix using L-GCN, $\phi$ is the activation function, $\mathbf{\hat{A}}_N$ is the modified node adjacency matrix, and $\mathbf{w}_2$, $\mathbf{w}_3$, $\mathbf{b}_2$, $\mathbf{b}_3$ are parameters in neural networks. Compared with GCN in Equation \eqref{Equation: GCN}, L-GCN utilizes deep neural networks to approximate the aggregation of link flows and incorporates node graph convolution.

\subsubsection{Inference with Historical O-D Matrix}
In this section, we present the structure of FL-GCN, which incorporates historical O-D flows. FL-GCN consists of two parts: link flows to O-D flows and historical O-D flows to predicted O-D flows. The evolution of link flows to O-D flows is denoted by L-GCN. We can use Convolutional Neural Networks (CNNs) or Fully Connected Networks (FCNs) to represent the evolution from historical O-D flows to the predicted O-D flows.

The predicted O-D matrix $\hat{\mathbf{x}}_h^1$ can be obtained by weighted summation of L-GCN outputs $\mathbf{x}{_h^L}$ and historical O-D flows $\mathbf{x}{_h^H}$,
\begin{align} \label{Equation: FL-GCN}
\hat{\mathbf{x}}_h^1 & = \psi \left( \mathbf{x}{_h^L} \mathbf{w_4} + \mathbf{x}{_h^H} \mathbf{w_5} + \mathbf{b}_4 \right),
\end{align}
where $\psi$ is the activation function, $\mathbf{w_4}$ and $\mathbf{w_5}$ are weighted parameters for two branches, and $\mathbf{b}_4$ is the parameter in neural networks.

The expression $\mathbf{x}{_h^H} \mathbf{w_5}$ in Equation \eqref{Equation: FL-GCN} is the structure of FCN, which denotes the nonlinear relationship from inputs to outputs. Since $\mathbf{x}{_h^H}$ is the $(n_d \times (n_d-1))$ vector of all historical O-D pairs, CNN can be used to capture adjacent O-D pair correlations. In this case, $\mathbf{x}{_h^H}$ is considered as an image with one channel. Then the FL-GCN with convolution is shown as:
\begin{align}\label{Equation: FL-GCN-CNN}
\hat{\mathbf{x}}_h^1 & = \psi \left( \mathbf{x}{_h^L} \mathbf{w_4} + \mathbf{x}{_h^H} \circ \mathbf{w_5} + \mathbf{b}_4 \right),
\end{align}
where `$\circ$' denotes the convolution operator.

\subsubsection{Objective Function}
The objective function shown in Equation \eqref{Equation: loss function} is to minimize the difference between the predicted and observed O-D flow,
\begin{align} \label{Equation: loss function}
\ell = \frac{1}{n_{od}} \sum_{i=1}^{n_{od}}\vert{x_i-\hat{x_i}}\vert,
\end{align}
where $x_i$ is the $i^{th}$ observed O-D flow, $\hat{x_i}$ is the $i^{th}$ predicted O-D flow, and $n_{od}$ is the number of O-D pairs in each interval.

\subsection{Deviation based Kalman Filter} 
\label{Section: Kalman_filter}
In Kalman filter, historical Origin-Destination (O-D) flows are firstly used to estimate initial state vector $\mathbf{\widetilde{x}}_0$, initial covariance $\mathbf{p}_0$, transition matrix $\mathbf{f}$ and transition error $\mathbf{w}$. Then we use observed link flows $\mathbf{y}$, assignment matrix $\mathbf{a}$ and measurement error $\mathbf{v}$ in time interval $h$ to predict O-D flows in interval $(h+k)$ during $k^{th}$ step prediction \cite{ashok2002estimation}.

Kalman filter consists of two steps: prediction and update. The state-space form in Equation \eqref{Equation1} can represent the spatial and temporal correlations. Since O-D flows in prior days incorporate similar patterns, deviations from historical data are considered to be the state-vector \cite{ashok2002estimation}.
  \begin{subequations} \label{Equation1}
    \begin{align} 
      &\mathbf{\widetilde{x}}_{h+1}-\mathbf{\widetilde{x}}{_{h+1}^H}=\sum_{p=h+1-q'}^{h}\mathbf{f}{_{h+1}^p} (\mathbf{\widetilde{x}}_p - \mathbf{\widetilde{x}}{_p^H}) + \mathbf{w}_{h+1}  \label{Equation1a}, \\
      &\mathbf{y}_h-\mathbf{y}{_h^H}=\sum_{p=h-p'}^h\mathbf{a}{_h^p} (\mathbf{\widetilde{x}}_p - \mathbf{\widetilde{x}}{_p^H}) + \mathbf{v}_h,          \label{Equation1b}
    \end{align} 
  \end{subequations}
where $\mathbf{\widetilde{x}}_{h+1}$ is the $(n_{od} \times 1)$ vector of O-D flows departing in interval $(h+1)$, and $\mathbf{\widetilde{x}}{_{h+1}^H}$ is the corresponding historical O-D flows. $\mathbf{f}{_{h+1}^p}$ is the matrix of time series effect of $( \mathbf{\widetilde{x}}_p - \mathbf{\widetilde{x}}{_p^H} )$ on $( \mathbf{\widetilde{x}}_{h+1}-\mathbf{\widetilde{x}}{_{h+1}^H} )$. $\mathbf{w}_{h+1}$ is the vector of transition errors. $\mathbf{a}{_h^p}$ is the assignment matrix which denotes the relationship between O-D flows and link traffic counts. $\mathbf{v}_h$ is the vector of measurement errors. $q'$ is the number of prior deviations, and $p'$ is the number of prior O-D intervals taken to calculate the link flows in $h^{th}$ interval.

Equation \eqref{Equation1a} is the auto-regressive progress, which denotes the temporal relationship among consequent O-D flows. Equation \eqref{Equation1b} represents the mapping from O-D flows to link traffic counts. The assignment matrix $\mathbf{a}{_h^p}$, which represents the nonlinear topology of transport networks, is mainly influenced by router choice behaviors and the mapping from path flows to link flows. Kalman filter uses prediction and update steps to minimize the estimation covariance, i.e., the prediction uncertainty.

Let $\partial{\mathbf{x}_h} = \mathbf{\widetilde{x}}_h-\mathbf{\widetilde{x}}{_h^H}$. To fit the structure of Kalman filter, we follow the technique of State Augmentation \cite{ashok1996estimation}, and re-define the state vector as:
\begin{align}
    & \mathbf{X_h}=[{\partial{\mathbf{x}_h}}^T \quad {\partial{\mathbf{x}_{h-1}}}^T \quad \ldots \quad {\partial{\mathbf{x}_{h-s}}}^T]^T,
\end{align}
in which $s=\max \left( p', q'-1 \right)$. The corresponding augmented transition matrix and error vector are:
  \begin{align}
    & \mathbf{F_h}=\left[\begin{matrix} 
    \mathbf{f}{_{h+1}^{h}}  & \mathbf{f}{_{h+1}^{h-1}} & \cdots   & \mathbf{f}{_{h+1}^{h-(s-1)}} & \mathbf{f}{_{h+1}^{h-s}} \\
    \mathbf{I}            & \mathbf{0}           & \cdots   & \mathbf{0}             & \mathbf{0}            \\
    \mathbf{0}            & \mathbf{I}           & \cdots   & \mathbf{0}             & \mathbf{0}            \\
    \vdots                & \vdots               & \ddots   & \vdots                 & \vdots                \\
    \mathbf{0}            & \mathbf{0}           & \cdots   & \mathbf{I}             & \mathbf{0} \\
  \end{matrix}\right]
  \end{align}
  and
  \begin{align}
     & \mathbf{W}_{h+1} = [{\mathbf{w}_{h+1}}^T \quad {\mathbf{0}}^T]^T.
  \end{align}
Then Equation \eqref{Equation1a} can be transformed into:
\begin{align} \label{x_aug}
\mathbf{X}_{h+1} = \mathbf{F}_h \mathbf{X}_{h} + \mathbf{W}_{h+1}.
\end{align}
For the state augmentation in measurement Equation \eqref{Equation1b}, the augmented link flows and assignment matrix are shown as:
\begin{align}
\mathbf{Y}_h = \mathbf{y}_h-\mathbf{y}{_h^H},
\end{align}
\begin{align}
\mathbf{A}_h = [\mathbf{a}{_h^h} \quad \mathbf{a}{_h^{h-1}} \quad \ldots \quad \mathbf{a}{_h^{h-s}}].
\end{align}
Then Equation \eqref{Equation1b} can be transformed into:
  \begin{align} \label{y_aug}
    \mathbf{Y}_h = \mathbf{A}_h \mathbf{X}_h + \mathbf{v}_h.
  \end{align}
Equation \eqref{x_aug} and \ref{y_aug} together can be used to predict the flows using the framework of Kalman filter, and $k^{th}$ step prediction can be realized by $k$ iterations using Equation \eqref{x_aug} \cite{ashok1996estimation}.

\subsection{Heterogeneous Prediction in Neural Networks and Kalman filter}
Silver et al. \cite{silver2016mastering} used a mixing parameter to balance node evaluation in deep neural networks and fast rollout policy. Motivated by this, a mixing parameter is used to balance heterogeneous prediction in graph neural networks and Kalman filter.
Deep neural networks are shown to be effective in recognizing spatial-temporal correlations using historical data. Kalman filter utilizes real-time link flows to update covariance matrix to minimize estimation uncertainty. Furthermore, Kalman filter uses the concept of deviation, which incorporates historical O-D information. A balancing weight $\lambda$ is used to reconcile the prediction in two methods. The final output is shown as:
\begin{align}
\mathbf{\hat{x}}_h = \lambda \mathbf{\hat{x}}{^1_h} + (1 - \lambda) \mathbf{\hat{x}}{_h^2},
\end{align}
where $\mathbf{\hat{x}}{^1_h}$ denotes the outputs of graph neural networks, and $\mathbf{\hat{x}}{_h^2}$ represents the outputs of Kalman filter. The prediction steps are shown as follows:


\begin{algorithm}[htb]
  \caption{Implementation steps in the combination model.}
  \label{algorithm:Framwork}
  \begin{algorithmic}[1]
    \Require
      Link adjacency matrix $\mathbf{A}_L$;
      
      Node adjacency matrix $\mathbf{A}_N$;
      
      Incidence matrix $\mathbf{P}$;
      
      Transition matrix $\mathbf{f}$;
      assignment matrix $\mathbf{a}$;
      
      Initial state vector $\mathbf{\widetilde{x}}_0$;
      initial covariance $\mathbf{p}_0$;
      
      Transition error $\mathbf{w}$;
      measurement error $\mathbf{v}$;
      
      Current link observations;
      
      Historical link observations;
      
      Historical O-D observations;
    \Ensure
      Predicted O-D flows;
    \State Predict O-D matrix $\mathbf{\hat{x}}{^1_h}$ using converged deep neural networks;
    
    \State Predict O-D matrix $\mathbf{\hat{x}}{^2_h}$ using estimated O-D matrix in prior intervals by Kalman filter;
    
    \State Update Kalman filter using link flows;

    \State Balance outputs in neural networks and Kalman filter using the balancing weight $\lambda$;
    \\
    \Return final O-D prediction $\mathbf{\hat{x}}_h$;
  \end{algorithmic}
\end{algorithm} 
\section{Case Study} \label{Section: case_study}
\subsection{Dataset}
\begin{figure*}[hbt]
  \centering
  \includegraphics[width=0.75\textwidth, trim=140 250 130 300,clip]{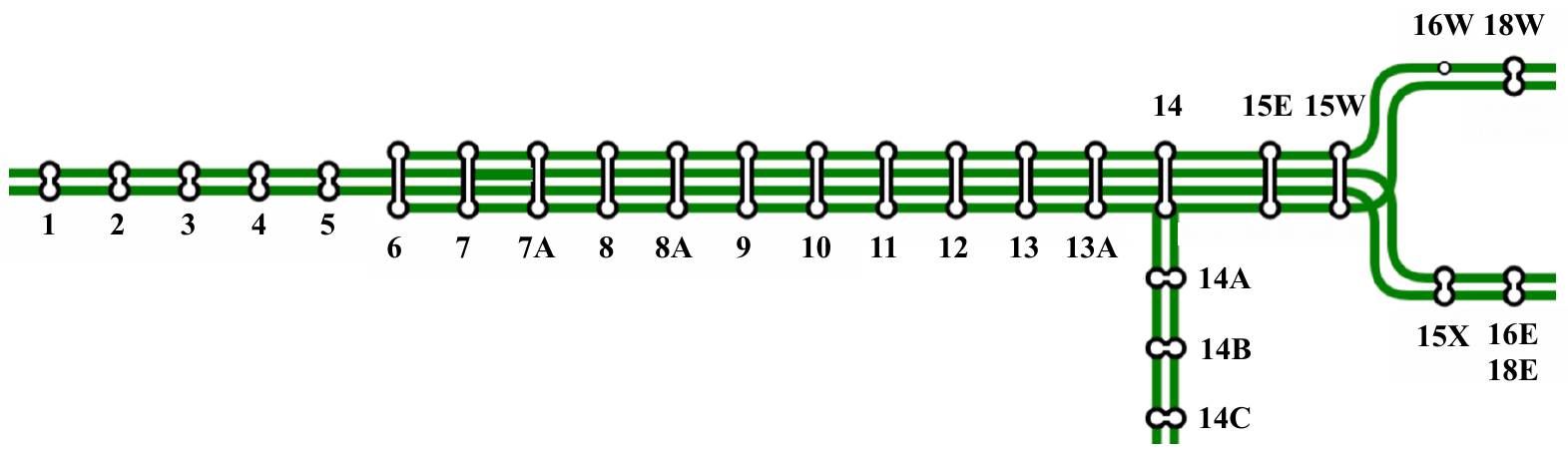}
  \caption{Section of NJ turnpike.}\label{fig:turnpike_map}
\end{figure*}

New Jersey (NJ) Turnpike data were used to test the applicability of our model combining Line Graph Convolutional Networks and Kalman filter for Origin-Destination (O-D) prediction. Figure \ref{fig:turnpike_map} shows the simplified map of NJ turnpike. The dataset provides anonymized entrance and exit times of each vehicle, which can be used to calculate aggregated O-D flows. The analysis period is from 6:15 A.M. to 9:45 A.M. with the length of departure interval being 15 minutes. The link flows are calculated at the entrance of each link. We assume that each vehicle has a constant speed $60$ mph to calculate aggregated link flows \cite{ashok1996estimation}. This assumption is unrealistic in real scenarios. However, all we need is a set of consistent O-D and link flows to implement our approach. The issue of whether the constant speed consumption is reasonable is not directly relevant. Since there is only one route for each O-D pair in the NJ Turnpike, the route choice effect is not considered.

There are 26 interchanges in this network (Figure \ref{fig:turnpike_map}). The O-D table is thus a $(26 \times 25)$ matrix. We assume that the entrance of each link is equipped with traffic sensors. Since there are two directions, the number of links is $50$. We use aggregated O-D flows and link flows from February 01 to May 31 in 2013, e.g., data in 4 months, to train the neural networks. Then we use O-D flows in June 2013, e.g., data in 1 month, to evaluate our prediction performance.

The platform for implementing our proposed algorithms is a server with 1 GPU (NVIDIA TITAN RTX, 24GB memory). When we trained the FL-GCN, 4 prior intervals, e.g., 1 hour data, were used to predict the next interval O-D flows. Historical O-D and link flows were from the same interval 7 days ago. Real-time and historical link flows were integrated into a $(50 \times 8)$ matrix. We constructed 3 layers of link Graph Convolutional Networks (GCNs) with respective filter size: $(8 \times 100), (100 \times 50)$, and $(50 \times 25)$, 1 layer of line graph transformation with filter size $(25 \times 25)$, and two layers of node GCNs with respective filter size: $(25 \times 50)$, and $(50 \times 25)$. The node adjacency matrix, link adjacency matrix, and incidence matrix were defined according to the topology shown in Figure \ref{fig:turnpike_map}. For the historical O-D demand part, we used Fully Connected Networks (FCNs) and Convolutional Neural Networks (CNNs) to evaluate the spatial correlations. The number of layers in FCN and the number of layers in CNN were both 3. The kernel size of CNN was $(3 \times 3)$. The size of $\mathbf{w_4}$ and the size of $\mathbf{w_5}$ in Equation \eqref{Equation: FL-GCN} were both $(25 \times 25)$.

In Kalman filter, we followed the assumptions in Ashok and Ben-Akiva \cite{ashok2000alternative}. Firstly, the structure of transition matrix remained constant over the whole day. Secondly, a flow between O-D pair $r$ for a period was related only to $r^{th}$ O-D flow of prior intervals, then the transition matrix $\mathbf{f}$ was diagonal. The assignment matrix $\mathbf{a}$ was calculated directly from link counts and O-D flows. In addition, we used deviations from the same interval 7 days ago as state variables \cite{ashok1996estimation}.

\subsection{Numerical Results}
In this section, we use three classical metrics to evaluate the prediction performance: Mean Absolute Error (MAE), Root Mean Square Error (RMSE) and Root Mean Square Error Normalized (RMSN), given by $MAE = \frac{1}{N} \sum_{i=1}^{N} \left| x{^N_i} - \hat{x}{^N_i} \right|$, $RMSE = \sqrt{\frac{1}{N} \sum_{i=1}^{N} \left( x{^N_i} - \hat{x}{^N_i} \right)^2}$, $RMSN = \frac{\sqrt{N \sum_{i} \left( x{^N_i} - \hat{x}{^N_i} \right)^2}}{\sum_{i} x{^N_i}}$, where $N$ is the total number of predicted O-D pairs in June 2013, $\hat{x}{^N_i}$ is the $i^{th}$ predicted O-D flow in the total O-D pairs, and $x{^N_i}$ is the corresponding ground truth. MAE and RMSE are used to measure the absolute difference, and RMSN is used to measure the relative difference.

We have conducted experiments under different prediction steps and  compared the performance of deep neural networks, Kalman filter and the mixing outputs. The models used are shown as follows:

(a). Historical: we use historical data in the same interval 7 days ago as the predicted flows.

(b). Kalman filter: deviation based Kalman filter shown in the methodology section.

(c). FL-GCN-FCN: this approach is FL-GCN without node convolution, and FCN is used to represent the evolution from historical O-D flows to predicted O-D flows.

(d). FL-GCN-CNN: this approach is FL-GCN without node convolution. CNN is used instead to denote the evolution from historical O-D flows to predicted O-D flows.

(e). LGCN-NGCN: this approach is FL-GCN with node convolution, where LGCN denotes the link graph convolution and NGCN represents the node graph convolution. In this approach, we use CNN to represent the evolution from historical O-D flows to predicted O-D flows.

(f). Mixing LGCN-NGCN: combining LGCN-NGCN with Kalman filter, and the parameter $\lambda$ is $0.8$.

(g). Mixing FL-GCN: combining FL-GCN-CNN with Kalman filter, and the parameter $\lambda$ is $0.8$.
 
The results in Table \ref{table:average_results} show that the mixing FL-GCN yields the best performance under different prediction steps. We first compare the performance between deep neural networks and Kalman filter. The results show that the errors in FL-GCN-FCN, FL-GCN-CNN, and LGCN-NGCN are smaller than those in Kalman filter and historical data, which indicate that deep neural networks are effective in recognizing dynamic spatial-temporal O-D patterns. The performance of deep neural networks becomes worse as we increase the prediction step.
 
\begin{table}[hbt]
  \begin{center}
  \caption{Prediction comparison among different models.}\label{table:average_results}
    \begin{tabular}{lllllll}
    \toprule
    Model                             & MAE           & RMSE           & RMSN                                              \\
    \midrule
     (a) 1-Step Predicted                                                                                                  \\
    Historical                        & 3.942         & 8.445          & 0.610                                             \\
    Kalman filter                     & 4.019         & 9.383          & 0.678                                             \\
    FL-GCN-FCN                        & 3.714         & 7.872          & 0.569                                             \\
    FL-GCN-CNN                        & 3.605         & 7.532          & 0.544                                             \\
    LGCN-NGCN                          & 3.607         & 7.557          & 0.546                                             \\
    Mixing LGCN-NGCN              & 3.594         & 7.486          & 0.541                                             \\
    Mixing FL-GCN                  & \textbf{3.585}         & \textbf{7.440}          & \textbf{0.537}                  \\
    \midrule
     (b) 2-Step Predicted \\
    Historical                        & 3.987         & 8.498         & 0.607                                              \\
    Kalman filter                     & 4.052         & 9.378         & 0.670                                              \\
    FL-GCN-FCN                        & 3.778         & 7.956         & 0.568                                              \\
    FL-GCN-CNN                        & 3.681         & 7.653         & 0.547                                              \\
    LGCN-NGCN                         & 3.684         & 7.663         & 0.547                                              \\    
    Mixing LGCN-NGCN               & 3.668        & 7.598         & 0.543                                               \\
    Mixing FL-GCN                 & \textbf{3.649}         & \textbf{7.557}         & \textbf{0.540}                   \\    
    \midrule
     (c) 3-Step Predicted \\
    Historical                        & 4.027         & 8.553         & 0.607                                              \\
    Kalman filter                     & 4.088         & 9.412         & 0.669                                              \\
    FL-GCN-FCN                        & 3.780         & 8.014         & 0.569                                              \\
    FL-GCN-CNN                        & 3.694         & 7.747         & 0.550                                              \\
    LGCN-NGCN                         & 3.743         & 7.820         & 0.555                                              \\    
    Mixing LGCN-NGCN              & 3.702         & 7.683         & 0.546                                              \\
    Mixing FL-GCN                  & \textbf{3.670}         & \textbf{7.642}         & \textbf{0.543}                   \\    
    \bottomrule
    \end{tabular}
    \end{center}
\end{table}

\begin{figure}[!ht]
  \centering
  \begin{subfigure}[b]{1.0\textwidth}
  \centering
  \includegraphics[width=0.64\textwidth, trim=60 110 60 140,clip]{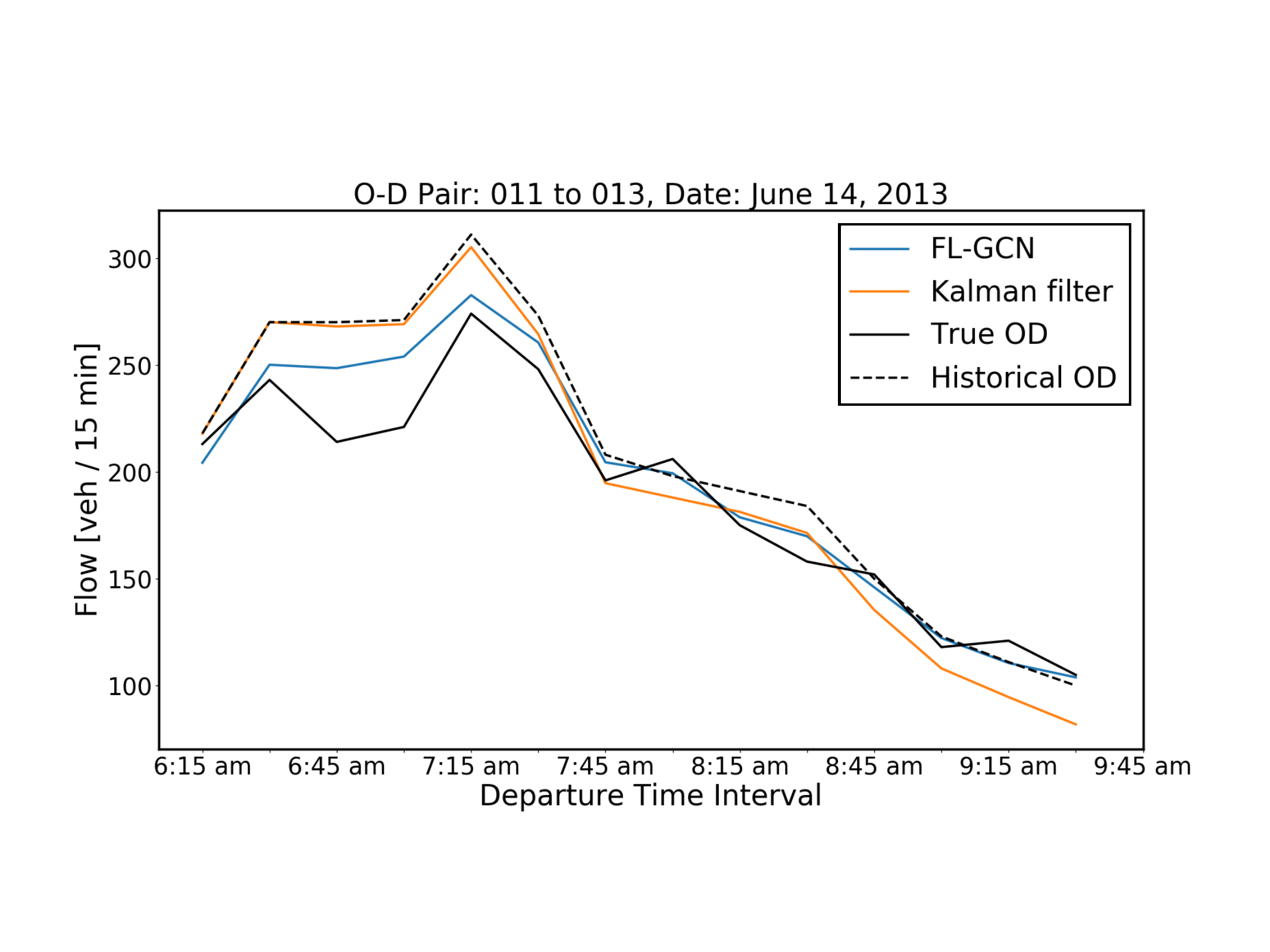}
  \caption{One step predicted}
  \label{fig:one_step_high_flow}
  \end{subfigure}
  \hfill
  
  \begin{subfigure}[b]{1.0\textwidth}
  \centering
  \includegraphics[width=0.64\textwidth, trim=60 110 60 140,clip]{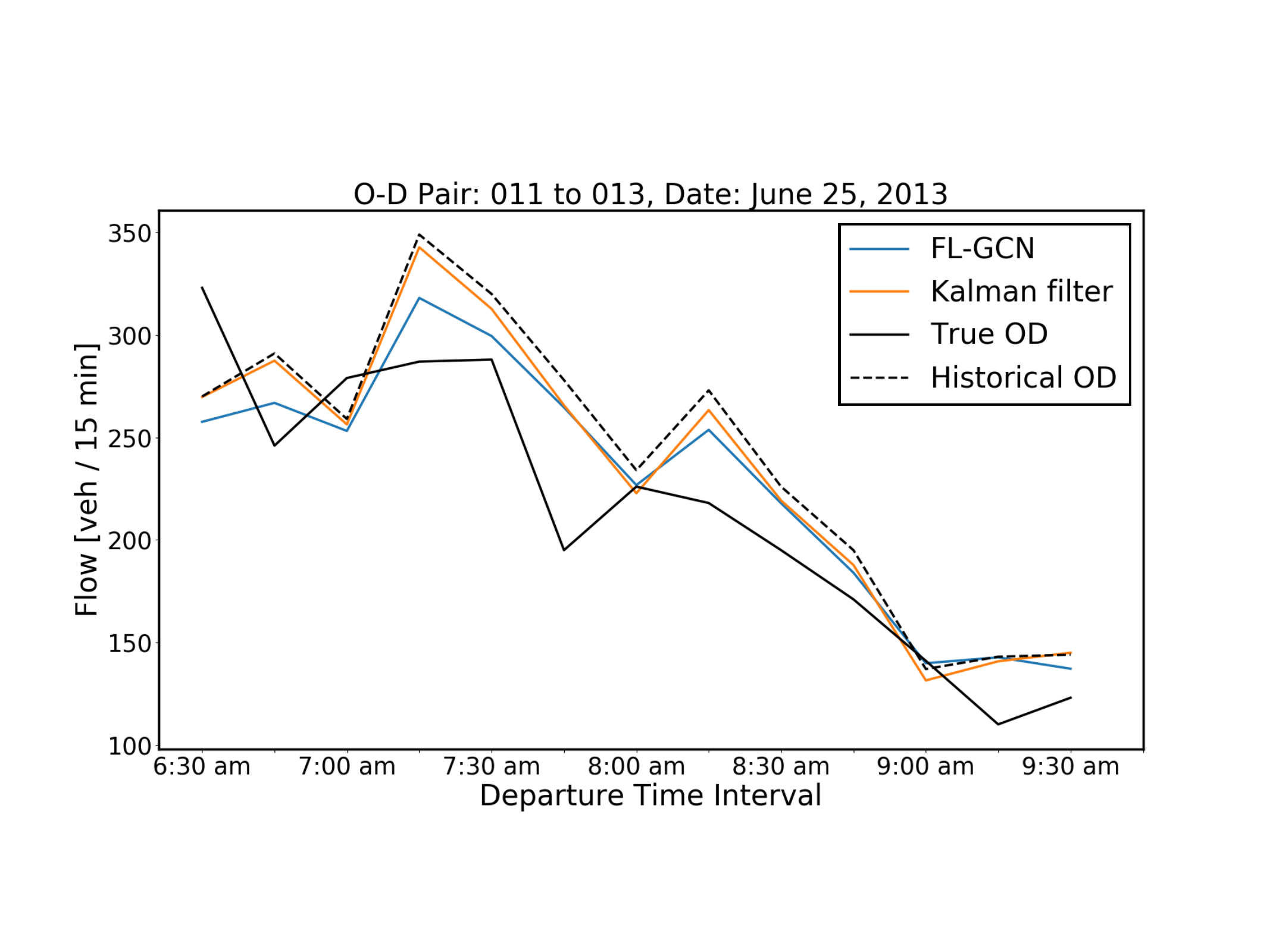}
  \caption{Two step predicted}
  \label{fig:two_step_high_flow}
  \end{subfigure}
  \hfill
  
  \begin{subfigure}[b]{1.0\textwidth}
  \centering
  \includegraphics[width=0.64\textwidth, trim=60 110 60 140,clip]{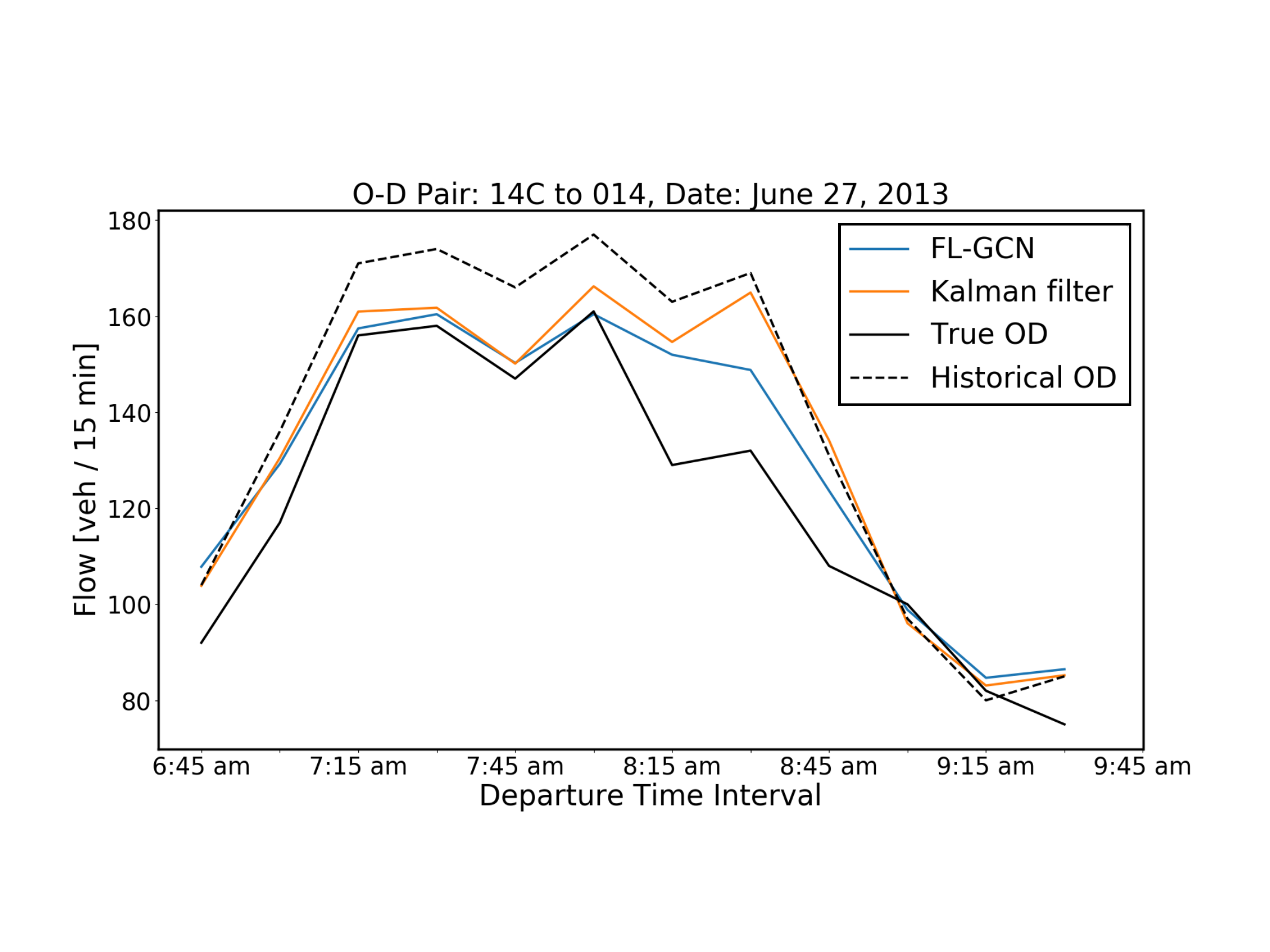}
  \caption{Three step predicted}
  \label{fig:three_step_high_flow}
  \end{subfigure}

  \caption{Model comparison under different prediction steps.} \label{fig:results_two_three_step}
\end{figure}

 We then randomly choose several O-D pairs to show the temporal prediction performance. The analysis period is the peak hour from 6:15 A.M. to 9:45 A.M. in June 2013. Figure \ref{fig:results_two_three_step} shows the results in one step, two step and three step predictions respectively. FL-GCN with CNN is used in these cases to yield better performance. From the comparison, we can see that FL-GCN can recognize the temporal patterns better than Kalman filter.
 
 Then we compare the performance using different neural network structures. The results in Table \ref{table:average_results} show that prediction errors in FL-GCN with CNN are smaller than those in FL-GCN with FCN, which indicates that adding CNN to recognize historical O-D correlations can improve prediction performance. The errors in LGCN-NGCN is almost the same with those in FL-GCN-CNN, which shows that adding node graph convolution is not necessary in flow information aggregation. FL-GCN with CNN yields the best performance compared with other models.
 
In addition, errors in mixing LGCN-NGCN are smaller than those in LGCN-NGCN, and errors in mixing FL-GCN are smaller than those in FL-GCN-CNN. The results show that adding appropriate Kalman filter into deep neural networks can improve prediction performance. Mixing FL-GCN is still better than mixing LGCN-NGCN due to better performance in FL-GCN with CNN.

In Table \ref{table:average_results}, the overall performance using different models are evaluated. The predictions of historical data are not much different from the performance of neural networks due to little variability in O-D flows between historical data and true values \cite{ashok1996estimation}. However, historical data cannot incorporate dynamic information, e.g., link flow variation 15 minutes ago, into predicted demands. In FL-GCN, more comprehensive data, i.e., $4$ prior intervals and historical O-D flows, are incorporated into neural networks.

Furthermore, the performance comparison with poor historical information, i.e., historical O-D flows are much different from true O-D flows, is shown in Table \ref{table:high_flows_poor_historical}. When the gap between historical O-D flow and true value is more than $100$ veh / $15 \min$, the O-D demand in that interval would be considered as the pair with poor historical information. The prediction comparison is under high flow level in Table \ref{table:high_flows_poor_historical}. The results show that all neural networks are significantly superior to historical data, especially the FL-GCN with CNN, when the poor historical values are used under the high flow level \cite{ashok1996estimation}.

\begin{table*}[!ht]
  \begin{center}
      \caption{Prediction comparison with poor historical information.}\label{table:high_flows_poor_historical}
    \begin{tabular}{ccccccc}
    
    \hline
    
    \multirow{2}*{Model}              &\multicolumn{3}{c}{Flows $\geq$ 100 [veh / $15 \min$] }                &\multicolumn{3}{c}{Flows $\geq$ 150 [veh / $15 \min$]} \\
    \cline{2-4}
    \cline{5-7}
                    & MAE   & RMSE   & RMSN         & MAE  & RMSE  & RMSN \\
    \hline
    (a) 1-Step Predicted    \\
    Historical      &143.685 &149.679 &0.799        &155.349 &161.396 &0.769 \\
    FL-GCN-FCN      &123.92  &132.357 &0.707        &130.634 &140.350 &0.668 \\
    FL-GCN-CNN      &113.868 &124.660 &0.666        &119.642 &131.784 &0.628 \\
    LGCN-NGCN       &122.085 &129.330 &0.691        &130.579 &138.069 &0.658 \\
    \hline
    (b) 2-Step Predicted \\
    Historical      &142.0     &147.494 &0.799        &153.555 &159.122 &0.767 \\
    FL-GCN-FCN      &114.090 &127.426 &0.690        &122.529 &136.673 &0.659 \\
    FL-GCN-CNN      &118.272 &126.936 &0.688        &125.888 &135.450 &0.653 \\
    LGCN-NGCN       &122.787 &129.427 &0.701        &133.299 &139.845 &0.674 \\
    \hline
    (c) 3-Step Predicted \\
    Historical      &140.197 &145.258 &0.799        &151.807 &156.967 &0.766 \\
    FL-GCN-FCN      &123.464 &131.827 &0.725        &131.247 &140.627 &0.686 \\
    FL-GCN-CNN      &117.356 &125.849 &0.692        &123.073 &132.894 &0.648 \\
    LGCN-NGCN       &121.617 &130.743 &0.719        &131.238 &141.056 &0.688 \\
    \hline
     \end{tabular}
     \end{center}
\end{table*}

\subsection{Analysis of the mixing weight}
In this section, the effect of mixing weight on prediction performance is analyzed. The results in Table \ref{table:high_flows} show the performance when flows are less than and more than $100$ veh / $15 \min$. Kalman filter yields better performance in the high flow level, especially when we use $RMSE$ and $RMSN$ as metrics. In Kalman filter, the predicted O-D flows are obtained by the summation of deviations and historical O-D flows. The pattern of O-D pair with higher flow is more significant than that with lower flow. Historical O-D flows have greater impact on the final prediction when the flow is high.

\begin{table*}[hbt]
  \begin{center}
      \caption{Prediction comparison under different flow levels.}\label{table:high_flows}
    \begin{tabular}{lccccccc}
    
    \hline
    
    \multirow{2}*{Model}              &\multicolumn{3}{c}{Flows $<$ 100 [veh / $15 \min$] }                &\multicolumn{3}{c}{Flows $\geq$ 100 [veh / $15 \min$]} \\
    \cline{2-4}
    \cline{5-7}
                                      & MAE           & RMSE           & RMSN           & MAE           & RMSE           & RMSN       \\
    \hline
    (a) 1-Step Predicted                                                                                                              \\
    Historical                        & 3.557         & 6.628          & 0.593          & 24.582        & 39.222 &0.250                       \\
    Kalman filter                     & 3.671         & 8.298          & 0.742          & 22.664        &33.395 &0.213                       \\
    FL-GCN-FCN                        & 3.348         & 6.219          & 0.556          & 23.262        &36.181 &0.230                 \\
    FL-GCN-CNN                        & 3.263         & 6.005          & 0.537          & 21.934        &34.117 &0.217                 \\
    LGCN-NGCN                         & 3.257         & 5.964          & 0.534          & 22.364        &34.793 &0.222     \\
    Mixing LGCN-NGCN               & 3.249         & 5.966          & 0.534          & 22.080        &33.926 &0.216       \\
    Mixing FL-GCN                   & 3.247         & 5.979          & 0.535          & 21.650        &33.258 &0.212                \\
    \hline
    (b) 2-Step Predicted \\
    Historical                        & 3.592         & 6.675          & 0.590          & 24.685        &39.040 &0.251              \\
    Kalman filter                     & 3.697         & 8.280          & 0.733          & 22.735        &33.246 &0.213               \\
    FL-GCN-FCN                        & 3.421         & 6.419          & 0.568          & 22.510        &34.980 &0.225                 \\
    FL-GCN-CNN                        & 3.332         & 6.088          & 0.538          & 21.994        &34.460 &0.221                 \\
    LGCN-NGCN                         & 3.326         & 6.041          & 0.534          & 22.502        &35.008 &0.225     \\
    Mixing LGCN-NGCN               & 3.315         & 6.051          & 0.535          & 22.232 &34.148 &0.219       \\
    Mixing FL-GCN                  & 3.304         & 6.057          & 0.536          & 21.740 &33.594 &0.216                \\
    \hline
    (c) 3-Step Predicted \\
    Historical                        & 3.625         & 6.722          & 0.591          & 24.803  &38.988  &0.253              \\
    Kalman filter                     & 3.724         & 8.296          & 0.729          & 22.900 &33.333 &0.216               \\
    FL-GCN-FCN                        & 3.394         & 6.280          & 0.552          & 23.729 &36.693 &0.238                 \\
    FL-GCN-CNN                        & 3.328         & 6.115          & 0.538          & 22.588 &35.073 &0.227                 \\
    LGCN-NGCN                         & 3.371         & 6.133          & 0.539          & 22.963 &35.755 &0.232     \\
    Mixing LGCN-NGCN               & 3.311         & 6.088          & 0.535          & 22.245 &34.096 &0.221       \\
    Mixing FL-GCN                  & 3.338         & 6.079          & 0.534          & 22.539 &34.652 &0.225                \\
    \hline
    \end{tabular}
    \end{center}
\end{table*}

The results in Table \ref{table:average_results} show that deep neural networks provide better performance than Kalman filter. We then use deep neural networks to output basic prediction and add proportional outputs of Kalman filter. Figure \ref{fig:results_two_three_step} shows that when FL-GCN is better than Kalman filter, there still exist some scenarios where the prediction errors in Kalman filter are less than those in deep neural networks, which indicates that Kalman filter can offset some worse predictions in FL-GCN. Combining both methods can increase prediction robustness to improve long-term performance.

\begin{figure}[!ht]
  \centering
  \begin{subfigure}[b]{1.0\textwidth}
  \centering
  \includegraphics[width=0.63\textwidth, trim=50 110 100 160,clip]{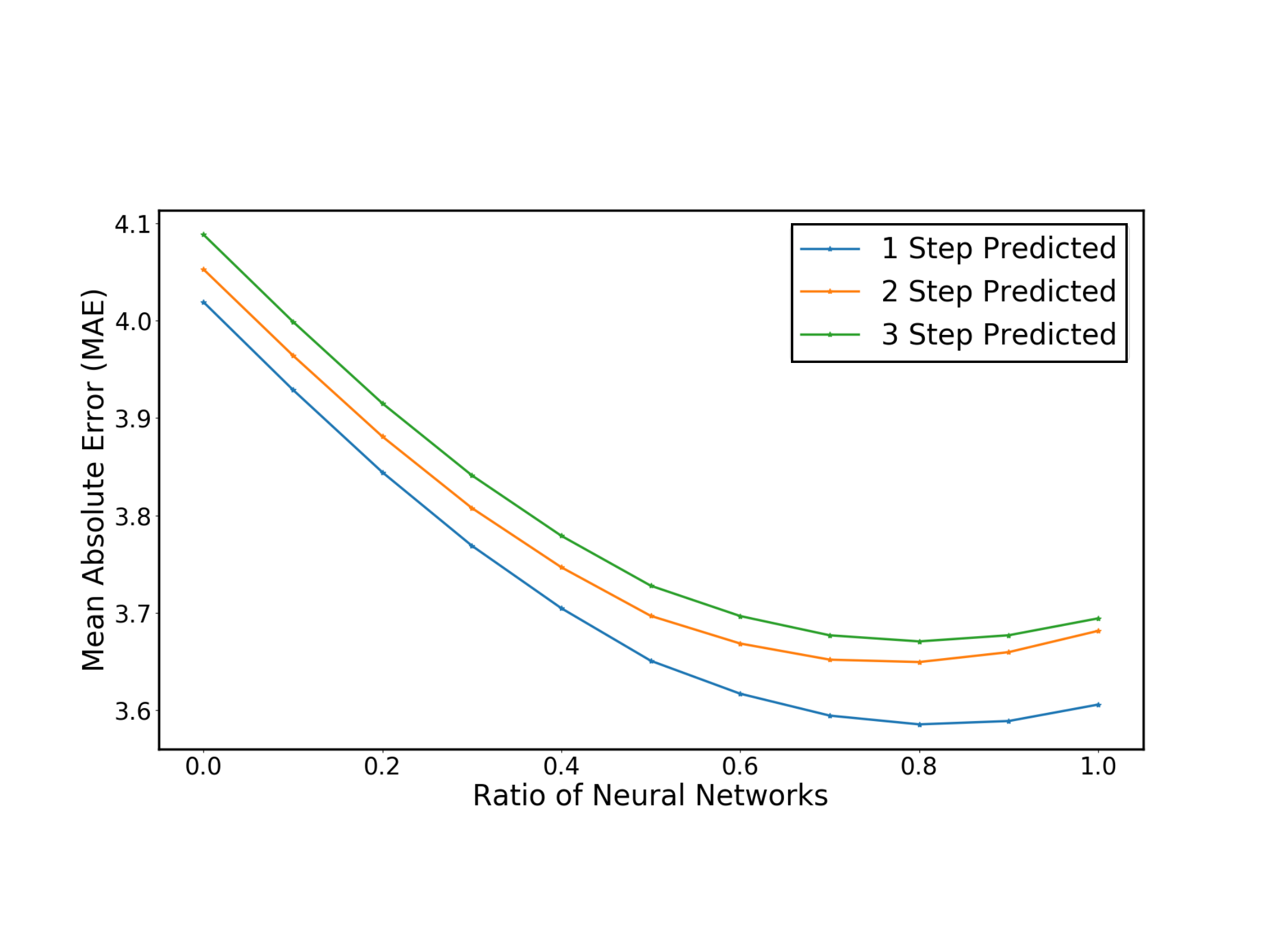}
  \caption{MAE performance}
  \end{subfigure}
  \hfill
  
  \begin{subfigure}[b]{1.0\textwidth}
  \centering
  \includegraphics[width=0.63\textwidth, trim=50 110 100 160,clip]{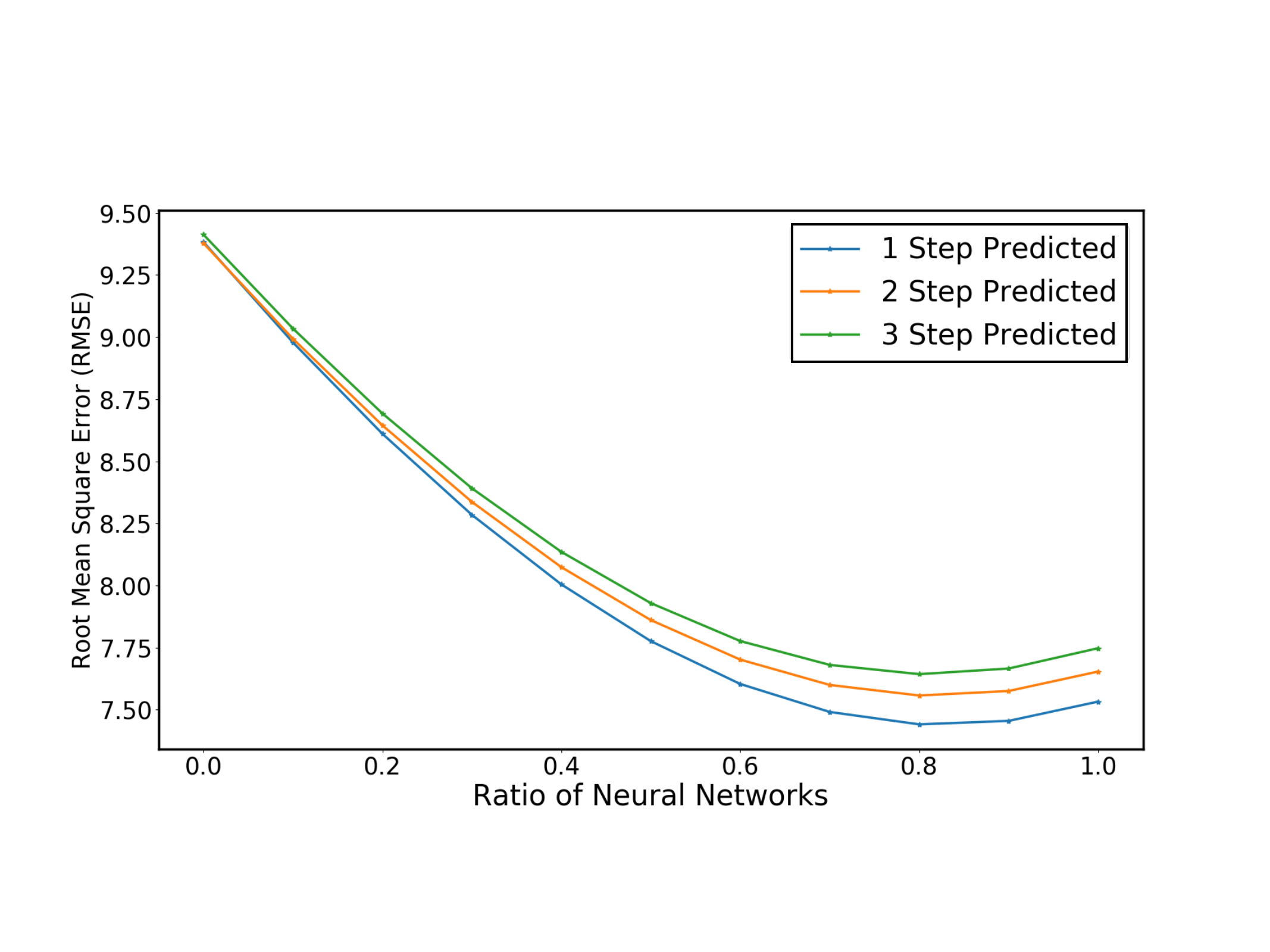}
  \caption{RMSE performance}
  \end{subfigure}
  \hfill
  
  \begin{subfigure}[b]{1.0\textwidth}
  \centering
  \includegraphics[width=0.63\textwidth, trim=50 110 100 160,clip]{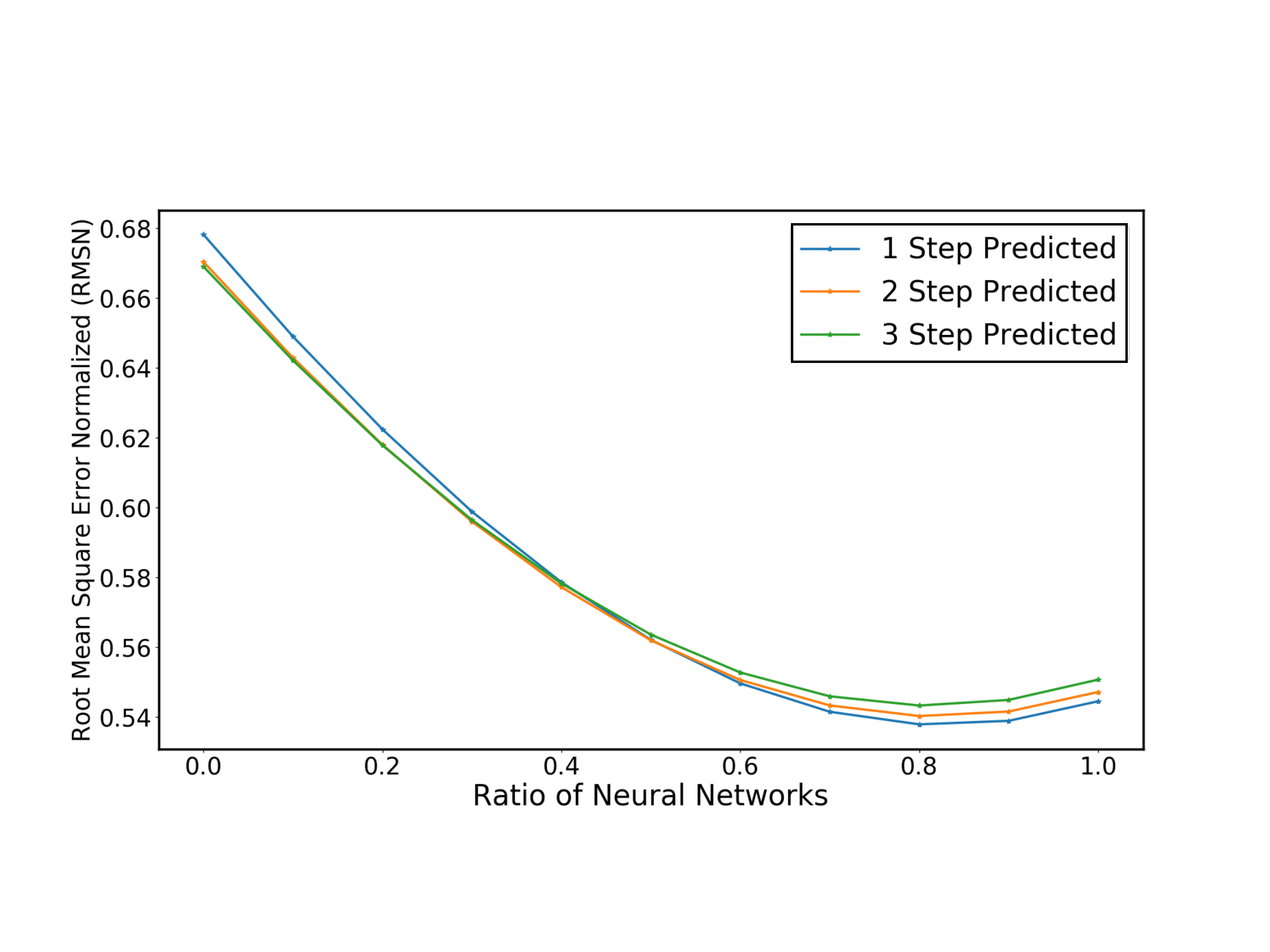}
  \caption{RMSN performance}
  \end{subfigure}
  
  \caption{MAE, RMSE and RMSN with ratio of neural networks.} \label{fig:rmse_rmsn_ratio}
\end{figure}

The mixing parameter is a key factor in determining the final outputs. Figure \ref{fig:rmse_rmsn_ratio} shows that MAE, RMSE and RMSN would change with the ratio of deep neural networks under three prediction steps. The results indicate that prediction errors would firstly decrease as we increase the weight of neural networks. The curves have the lowest prediction errors when the ratio is 0.8. After that, prediction errors would increase as we add Kalman filter. The results show that deep neural networks play a major part in predicting O-D flows, and adding proportional outputs of Kalman filter could improve prediction performance.

\subsection{Interpretation of Graph Neural Networks}
\begin{figure}[!ht]
  \centering
  \begin{subfigure}[b]{1.0\textwidth}
  \centering
  \includegraphics[width=0.63\textwidth, trim=90 120 170 170,clip]{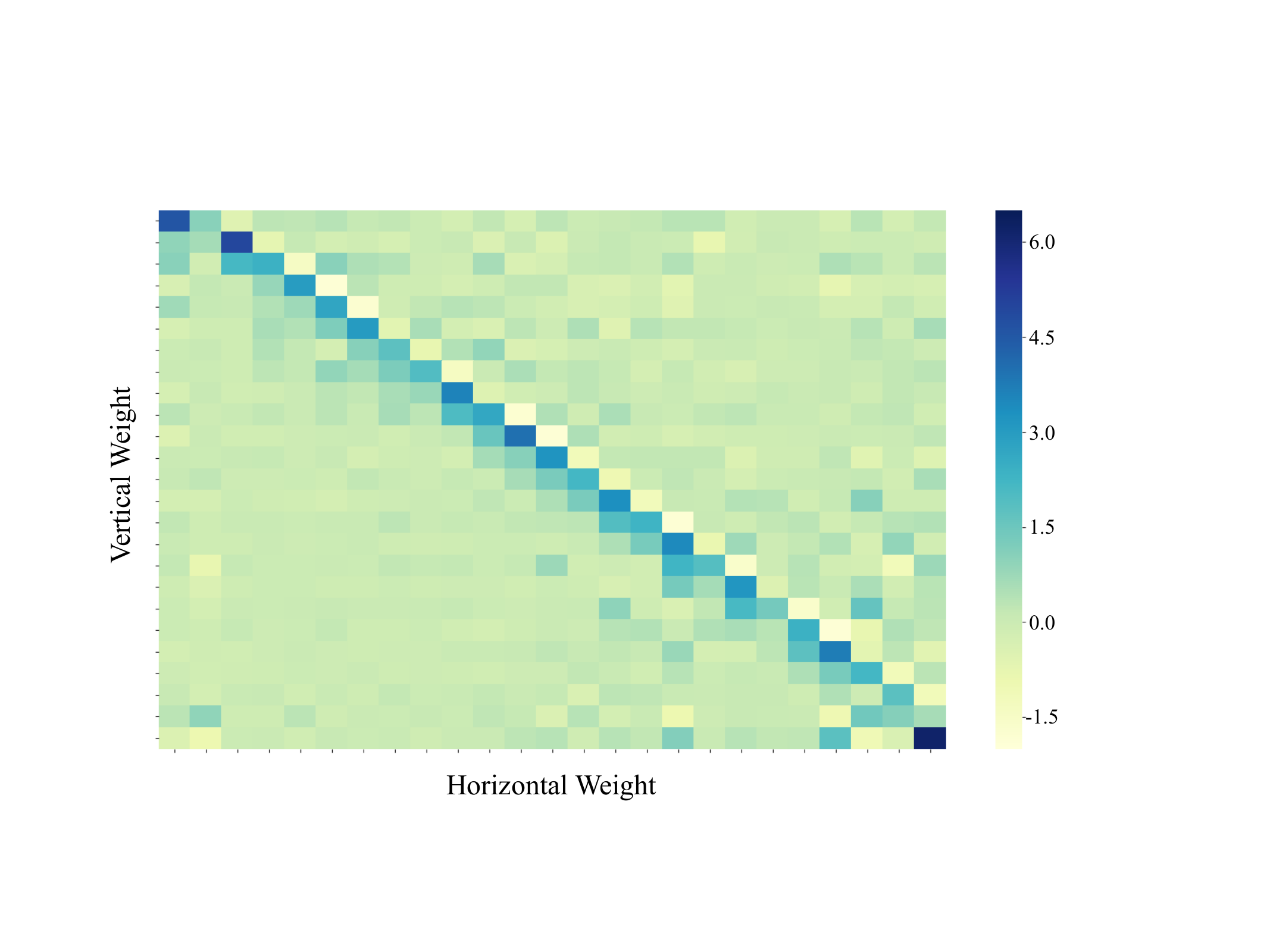}
  \caption{Historical O-D weight}
  \label{fig:weight_his_OD}
  \end{subfigure}
  \hfill
  
  \begin{subfigure}[b]{1.0\textwidth}
  \centering
  \includegraphics[width=0.63\textwidth, trim=90 120 170 170,clip]{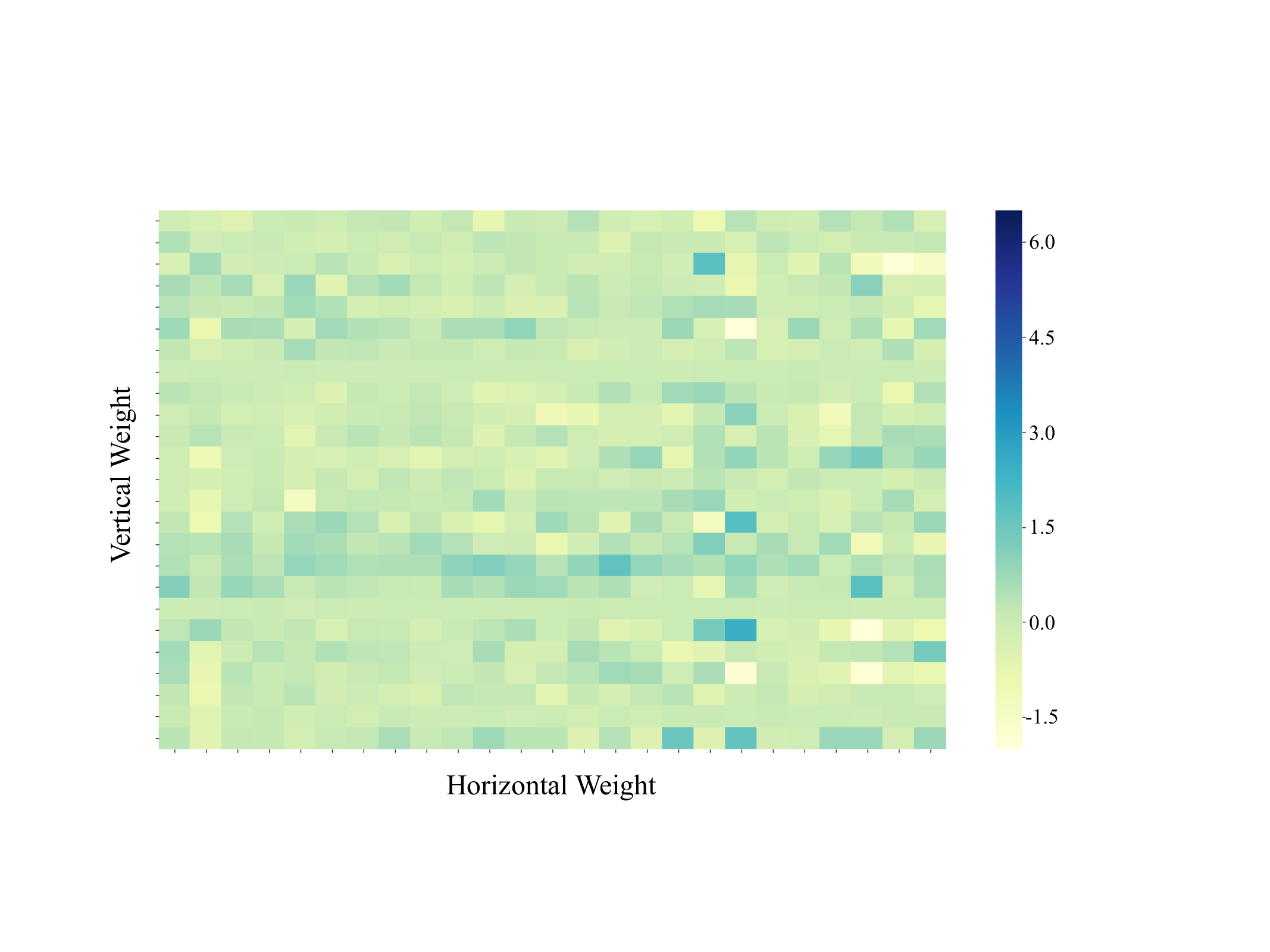}
  \caption{Link flow weight}
  \label{fig:weight_link}
  \end{subfigure}
  
  \caption{Converged weights in convolutional neural networks  and graph neural networks.} \label{fig:weight}
\end{figure}

In this section, we visualize and analyze the weights in converged deep neural networks \cite{zeiler2014visualizing, yosinski2015understanding}. In Equation \eqref{Equation: FL-GCN-CNN}, $\mathbf{w}_4$ denotes the weight in line graph convolution networks, and $\mathbf{w}_5$ represents the weight in historical O-D flows. We visualize the converged weights to show the effect of link flows and historical O-D flows. Figure \ref{fig:weight_his_OD} denotes the weight of historical O-D flows, and Figure \ref{fig:weight_link} shows the weight of link graph outputs. The results in the heat map show that historical weight $\mathbf{w}_5$ is larger than link weight $\mathbf{w}_4$, and the pattern is more significant in the diagonal direction, which indicates that historical O-D flows have greater impact than link flows on the final prediction.

In Figure \ref{fig:results_two_three_step}, we can see that both FL-GCN and Kalman filter are inferred based on historical O-D flows and updated by link flows. FL-GCN incorporates historical O-D flows into deep neural networks, and utilizes CNN to recognize spatial correlations among O-D pairs. The prediction of Kalman filter is based on deviation from historical O-D flows, which also emphasizes the effect of historical O-D flows. 
\section{Conclusions} \label{Section: conclusion}
This paper proposes a new framework that combines graph neural networks and Kalman filter to predict Origin-Destination (O-D) demands along a closed highway. In graph neural networks, we design the novel Fusion Line Graph Neural Networks (FL-GCN) including link graph convolution and node graph convolution, which provides a general deep learning frameworks that can be used to deal with problems related to spatial-temporal aggregation from links to nodes. We use New Jersey Turnpike data to evaluate the performance of our model. The results show that our model 
that combines FL-GCN with Kalman filter yields the best performance in recognizing traffic spatial-temporal patterns. We also analyze the heterogeneous prediction of neural networks and Kalman filter, the results show that prediction errors obtain the minimum values when the ratio of deep neural networks approximates 0.8. In addition, we visualize the converged weights to understand the deep neural network. The results show that historical O-D flows have greater weights and impacts than link weights in the structure of FL-GCN.

This work can be extended in several directions. First, we can use the proposed approach to deal with missing observations due to sensor failures. Second, we can extend the approach to recurrent line graph neural networks to recognize time-series patterns. Third, we can generalize the proposed framework by adding more traffic information (e.g. speed).

\bibliographystyle{IEEEtran}
\bibliography{od_prediction}   
\end{document}